\documentclass[10pt,twocolumn,letterpaper]{article}

\usepackage{cvpr}              

\usepackage[dvipsnames]{xcolor}

\definecolor{cvprblue}{rgb}{0.21,0.49,0.74}
\usepackage[pagebackref,breaklinks,colorlinks,citecolor=cvprblue]{hyperref}
\usepackage{lipsum}
\usepackage{adjustbox}
\usepackage{amssymb}
\usepackage{pifont}
\usepackage{paralist}
\usepackage{bm}
\usepackage{comment}

\def\paperID{00029} 
\def\confName{CVPR}
\def\confYear{2024}

\title{Watching Swarm Dynamics from Above: \\ A Framework for Advanced Object Tracking in Drone Videos}

\author{Duc Pham\textsuperscript{1}\ , Matthew Hansen\textsuperscript{3}\ , Félicie Dhellemmens\textsuperscript{2}, Jens Krause\textsuperscript{3,4}, Pia Bideau\textsuperscript{2,4}\thanks{Corresponding Author: pia.bideau@inria.fr}\footnotemark[1]\\
\textsuperscript{1}Technical University Berlin, \textsuperscript{2}Univ. Grenoble Alpes, Inria, CNRS, Grenoble INP, LJK, \\ \textsuperscript{3}Leibniz Institute for Freshwater Ecology and Inland Fisheries, Berlin\\
\textsuperscript{4}Science of Intelligence, Research Cluster of Excellence, Berlin
}

\begin{document}
\maketitle
\begin{abstract}
Easily accessible sensors, like drones with diverse onboard sensors, have greatly expanded studying animal behavior in natural environments. Yet, analyzing vast, unlabeled video data, often spanning hours, remains a challenge for machine learning, especially in computer vision. Existing approaches often analyze only a few frames. Our focus is on long-term animal behavior analysis.
To address this challenge, we utilize classical probabilistic methods for state estimation, such as particle filtering. By incorporating recent advancements in semantic object segmentation, we enable continuous tracking of rapidly evolving object formations, even in scenarios with limited data availability.
Particle filters offer a provably optimal algorithmic structure for recursively adding new incoming information. We propose a novel approach for tracking schools of fish in the open ocean from drone videos. Our framework not only performs classical object tracking in 2D, instead it tracks the position and spatial expansion of the fish school in world coordinates by fusing video data and the drone's on board sensor information (GPS and IMU). 
The presented framework for the first time allows researchers to study collective behavior of fish schools in its natural social and environmental context in a non-invasive and scalable way.
\end{abstract}    
\section{Introduction}
\label{sec:intro}

\begin{figure}
    \centering
    \includegraphics[width=\linewidth]{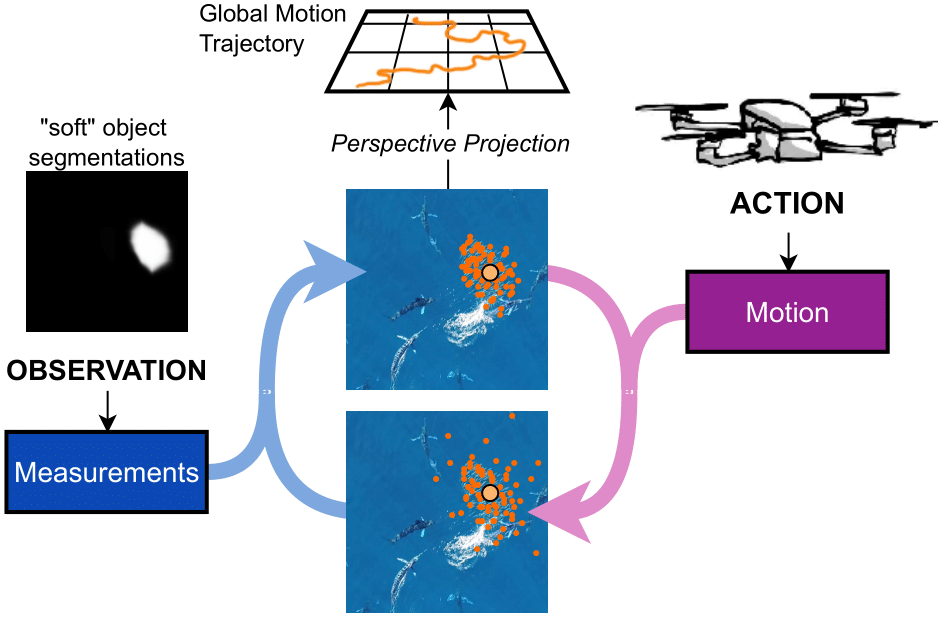}
    \vspace{-0.5ex}
    \caption{\textbf{Swarm Dynamics from Above (SwDA),} a framework for tracking collective behavior from drone videos. The recursive architecture of a particle filter, coupled with frame-by-frame semantic segmentation allows tracking over long time horizons.
    }
    \label{fig:overview}
    \vspace{-4mm}
\end{figure}

\begin{table*}
\centering
\small
\begin{adjustbox}{max width=0.95\linewidth}
\setlength{\tabcolsep}{3pt}
\begin{tabular}{lc ccc ccc}
\toprule 
Method & Object (Animal) & Animal Group & Moving & 2D & 3D & Temporal & Tracking \\
& Tracker & Representation & Camera & Trajectory & Trajectory & Scope & Environmnent\\
\midrule
DeepLabCut~\cite{mathis2018deeplabcut} & Point/Skeleton & multiple individuals & \ding{51} & \ding{51} & \ding{55} & long-term & terrestrial/marine\\ 
Particle Video~\cite{harley2022} & Point & - & \ding{51} & \ding{51} & \ding{55} & short-term & terrestrial\\ 
Follow Anything~\cite{maalouf2024follow} & Point/Shape & - & \ding{51} & \ding{51} & \ding{55} & long-term & terrestrial/marine \\
Quantifying Behavior with a Drone~\cite{koger2022} & Point/Bounding Box & multiple individuals & \ding{51} & \ding{51} & \ding{51} & long-term & terrestrial \\
Swarm Dynamics from Above (SwDA) & Point/Shape & swarm & \ding{51} & \ding{51} & \ding{51} & long-term & marine \\
\bottomrule
\end{tabular}
\end{adjustbox}
\vspace{-0.5ex}
\caption{{\bf Literature review} on approaches for tracking animal behavior captured by a moving camera. Methods deviate in their object representation (point/skeleton/bounding box), non of existing methods directly track a pixel accurate object shape over long time horizons. SwAD is the only method designed for marine environments.
}
\label{table:literature_review}
\vspace{-0.3cm}
\end{table*}

Schools of fish, flocks or birds or herds of sheep - in nature, collectives exhibit remarkable behaviors as they seemingly synchronize their movements within the group. Flexibly the collective changes their formation, density, speed and may even change their formation completely by smoothly dividing or merging~\cite{sumpter2006principles}. These phenomena lead to marvellous visual spectacles that we can observe in the sky or in water. Both - aerial and marine environments,  allow researchers to study collective behavior and the interaction between multiple collectives in natural environments that are minimally influenced by constraints imposed by the terrestrial structure. Therefore these ecological environments present an unique opportunity to study behavior of animal groups consisting of several thousand individuals~\cite{hansen2022mechanisms}. 
The complexity of such group-motions go far beyond patterns that can be described with traditional representations like skeletons, key-points or bounding boxes mimicking the spatial configuration of an object~\cite{hughey2018challenges}.
Latest research mostly focused on tracking behavior of one or more individuals~\cite{mathis2018deeplabcut, harley2022, maalouf2024follow, koger2022, walter2021trex, graving2019deepposekit, pereira2022sleap, nath2019using, perez2014idtracker}. In this work, we tackle the problem of tracking collective behavior. The analysis of collective behavior shifts the perspective from previously specific observations of individual behavior to a global view encompassing group dynamics involving multiple individuals, often of the same species. 
Our specific goal is to track a school of fish as a cohesive group, capturing key features like position, spatial extent, and shape of the entire collective. 
We propose a method that leverages classical probabilistic state estimation via particle filtering by integrating advances in semantic object segmentation.
The particle filter allows for recursively adding new incoming information over long time horizons. We continuously track the fish school while integrating learned segmentation masks and the drone's movement (GPS and IMU).
Our contributions can be summarized as follows:
\begin{compactenum}[i.]
    \item For the first time, an approach is introduced that enables the analysis of animal behavior in real-world coordinates, even in demanding environments lacking geographical structures or unique landmarks. 
    \item The proposed methodology combines learning-based approaches for semantic segmentation with recursive Bayesian filtering, resulting in high performance even in low-data regimes.
    \item Its temporal robustness is demonstrated in the tasks of trajectory and shape estimation. Both of which are evaluated over extended time periods ($\sim$5 minutes).
    \item Code and the new dataset for tracking collective animal behavior will be released a long an extended version of this workshop paper.
\end{compactenum}

\section{Related Work}
Algorithms have been developed for tracking animal behavior in well-controlled laboratory experiments using stationary cameras~\cite{dell2014automated,mathis2018deeplabcut}, but also for handling challenging field experiments~\cite{kays2015terrestrial,nathan2022big, koger2022,maalouf2024follow}. While the ultimate goal is to analyze movements of single animals or groups in the wild, the natural environment significantly determines the capability of extracting accurate information about the animal's movement. Table~\ref{table:literature_review} provides an exemplary overview of the diversity of existing techniques and their applications. Techniques range from high-precision tracking in structured environments to more adaptable methods suited for the complex and variable conditions like marine environments. In terrestrial environments well-identifiable landmarks help to reconstruct the 3D trajectory from video sources~\cite{haalck2023cater,koger2022}. The ability to extract 3D behavioral trajectories significantly degrades with the opportunity of identifying high-quality landmarks. Estimating 3D trajectories in marine environments has not been widely addressed~\cite{matley2022global,maalouf2024follow}. The primary reasons are the lack of geometric structure needed to reconstruct 3D movement trajectories, making it a challenging vision and robotics problem. Additionally, the accessibility of animals found far off the coast makes the use of technical communication devices that rely on a fixed base (receiver) impractical. This work for the first time processes visual data (videos) and GPS/IMU information recorded by a drone to extract real world trajectories of a fish school in the Pacific Ocean 10-30 km offshore Baja California Sur.
\section{Visual Tracking of Swarm Dynamics}
\label{sec:method}
We present a novel framework for tracking swarm dynamics from drone recordings. In Section~\ref{sec:background} we introduce the recursive structure of the particle filter that unifies accurate frame-wise object detections and the drone's sensor data. Building upon the algorithmic stracture of the particle filter, Section~\ref{sec:swda} introduces \textit{Swarm Dynamics from Above}. 

\subsection{Background: Particle Filter}
\label{sec:background}
We examine the task of inferring a hidden state $s$ from a sequence of observations $o$ and performed actions $a$, i.e. the probability distribution of $p(s_{t}| o_{0:t}, a_{0:t}) = \text{bel}\big(s_t\big)$. In other words, we wish to localize the object in world coordinates from frame-wise object detections and the drone's pose provided by its GPS and IMU data.
Following the principle of Bayesian filtering, 
a belief over the object's location evolves over time through two sequential steps that iterativly track the object's location: 1) prediction using action $a_t$ and 2) update using observation $o_t$:
\begin{align}
\label{eq:marginalization}
\overline{\text{bel}}\big(s_t\big) &= \int p\big(s_t| s_{t-1}, a_t\big) \text{bel}\big(s_{t-1}\big)\,d s_{t-1}\\
\text{bel}\big(s_t\big) &= \eta p\big(o_t|s_t\big) \overline{\text{bel}}\big(s_t\big)
\end{align}
The Bayes filter computes $\text{bel}\big(s_t\big)$ recursively from $\text{bel}\big(s_{t-1}\big)$ while incorporating the new information contained in $a_t$ and $o_t$. 
\textit{How to represent this believe?} Particle filters are a way to efficiently represent an arbitrary (non-Gaussian) distribution. In case of a particle filter a set of particles $s_t^{[0]}, ..., s_t^{[N]}$ and weights $w_t^{[0]},..., w_t^{[N]}$ serve as an approximation to the probability distribution to be estimated. More specifically particles are iteratively moved, weighted and resampled.
The particle filter implements the prediction step by moving each particle stochastically, which is achieved by sampling from a generative motion model, 
$s_t^{[i]}\sim p(s_t^{[i]} | a_t, s_{t-1}^{[i]})$.
During the measurement update the weight of each particle $i$ is set to the observation likelihood,
$w_t^{[i]} =  p(o_t | s_{t}^{[i]})$
and particles are resampled accordingly. Following the underlying recursive structure of the particle filter we introduce \textit{Swarm Dynamics from Above}, a model that allows robust tracking of animal behavior in real world-coordinates, even in demanding environments that lack geometric structures or unique landmarks. In these highly demanding environments any classical tracking algorithms that solely rely on motion estimates through optical flow are typically prone to errors~\cite{horn1986robot, derpanis2012dynamic}.

\subsection{Tracking of Swarm Dynamics in Drone Videos}
\label{sec:swda}
The drone is equipped with a high resolution camera, gimbal for image stabilization and on board sensors - IMU and GPS, that provide the absolute drone pose (in geographic coordinates), its translational velocity measured in m/s and its rotation (pitch, yaw and roll) in degree. Measurements are converted into cartesian coordinates with its origin being at the position of the drone projected onto the ground. 
The camera's pose is determined from provided sensor measurements using a standard Kalman filtering approach. Given an accurate estimate of the drone's pose the following paragraphs outline the particle tracking framework with its motion model and measurement model that keep track over the objects position on the image plane over long time horizons. At time $t=0$ the particle filter is initialized with a uniformly distributed set of $P$ particles. Section~\ref{sec:frame2world} summarizes the conversion of the 2D trajectory into a global motion trajectory. It's worth noting that 2D tracking encompasses information about the drone's movement. 

\begin{figure}
    \centering
    \includegraphics[width=0.6\linewidth]{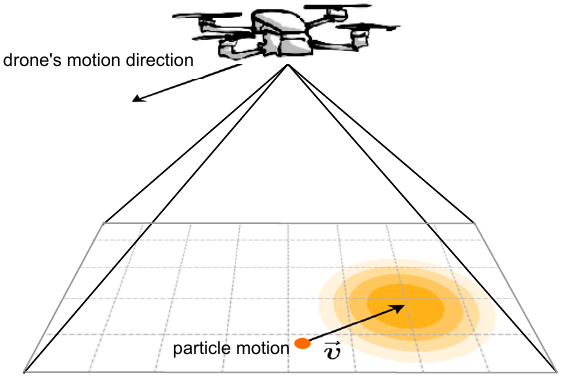}
    \vspace{-0.5ex}
    \caption[Optical flow compensation illustration]{{\bf Illustration of the motion model.} Each particle is displaced by the induced motion vector due to the drone's movement}
    \label{fig:method:optical_flow}
    \vspace{-4mm}
\end{figure}

{\bf Motion Model.} The complex open ocean environment, lacking clear geometrical structures and often featuring sun reflections due to wind or animal movements near the water surface, hinders direct motion estimation from optical flow. Optical flow relies on the brightness consistency assumption~\cite{horn1981determining}, which doesn't hold aforementioned cases. Instead of estimating particle movement from video frames, we infer the particle movement induced by the camera's motion. 
Let the camera rotation be defined by $[A, B, C]$ and its translation by [U, V, W], following rules of perspective projection the flow can be geometrically determined via:
\small
\begin{align}
    \label{eq:flow}
    \vec{v} =& \frac{1}{z} \begin{bmatrix} -fU + xW \\ -fV + yW \end{bmatrix} + \begin{bmatrix} \frac{A}{f}xy - Bf - \frac{B}{f}x^{2} + Cy\\ Af + \frac{A}{f}y^{2} - \frac{B}{f}xy - Cx\end{bmatrix},
\end{align}
\normalsize
here $f$ is the camera's focal length in pixels and $z$ the camera's height (distance to the captured scene)~\cite{horn1986robot}. Each particle is displaced by $\vec{v}$, and Gaussian noise is introduced by resampling from a Gaussian distribution with the mean equal to the updated particle position (Fig.~\ref{fig:method:optical_flow}). 

{\bf Measurement Model.} Incoming new observations $o_t$ continuously update the belief over the object's pose using Bayes’ rule.
We estimate \textit{soft} segmentation masks $o_t$ by first training DeeplabV3~\cite{chen2017rethinking} with a MobileNet backbone architecture~\cite{howard2017mobilenets} on the new dataset for swarm tracking. For training we minimize the binary cross-entropy loss, which is equivalent to maximizing the likelihood of the data. This process yields soft segmentation masks obtained during network inference. Particles are weighted according to estimated segmentation masks and resampled accordingly. We utilize roulette wheel sampling for particle resampling, a technique that incorporates elements from evolutionary computation methodologies~\cite{golberg1989genetic}.

Starting from initially uniformly distributed particles, particles quickly adapt to the swarm's shape and are capable to capture its dynamics - displacement and shape deformation. To achieve this goal, we introduced a model that iteratively conducts \textit{prediction} by considering the drone's actions and then \textit{updates} the prediction using learned object segmentation masks (Fig.~\ref{fig:overview}). This improves tracking accuracy, particularly in low data regimes, a prevalent challenge frequently encountered in animal behavior research.

\subsubsection{Estimating the Global Movement Trajectories}
\label{sec:frame2world}
Global movement trajectories are computed following the physical principles of image formation. Given the camera's pose, the camera's extrinsic parameters $R$ and $\bm{t}$ determining the 3D camera motion can be determined. Intrinsic camera parameters $K$ such as focal length and principle point offset are known. Therefore the global movement trajectory can be obtained as follows: $\bm{p}=K[R|\bm{t}]\bm{P}$, where $\bm{P}$ denotes the world point and $\bm{p}$ denotes its projection onto the image plane~\cite{hartley2003multiple}.
We define the origin of the global coordinate system to be at the initial drone pose projected onto the ground. Furthermore it is assumed that the camera's height is equal to the distance to the object being tracked, implying that the swarm is situated at the water surface. 

\section{Experiments}
\label{sec:exp}

We present novel data for tracking the dynamics of a school of fish being pursued by predators. The strong interaction between predator and prey results in highly expressive and unique swarm dynamics. We include an overview of the experimental setup, evaluation metrics covering tracking, shape segmentation accuracy, and the reconstruction of world coordinates from image detections.

{\bf Implementation Details.} Data was recorded with the drone DJI Phantom 4.
For Network training we utilized DeeplabV3 pretrained on ImageNet~\cite{deng2009imagenet} and further train using the BCEWithLogitLoss and AdamW optimizer. The networks learning rate was set to 1e-3. The network was trained for 50 epochs and the model that reaches lowest error on validation set was chosen. During particle tracking the objects shape was approximated with 1000 particles.

\begin{figure*}
\centering
\begin{subfigure}[b]{.27\textwidth}
\centering\includegraphics[width=\textwidth]{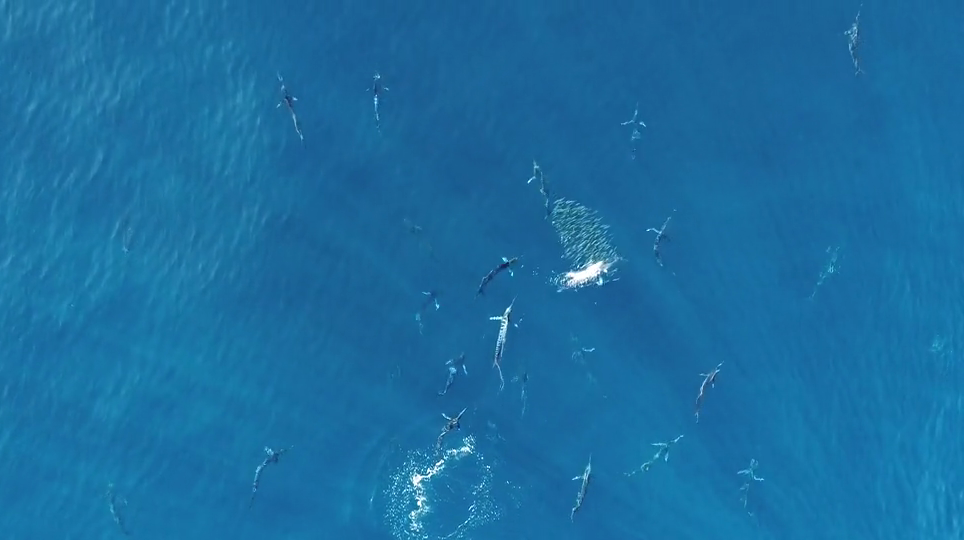}
\end{subfigure}\hspace{1em}\vspace{0.5em}
\begin{subfigure}[b]{.27\textwidth}
\centering\includegraphics[width=\textwidth]{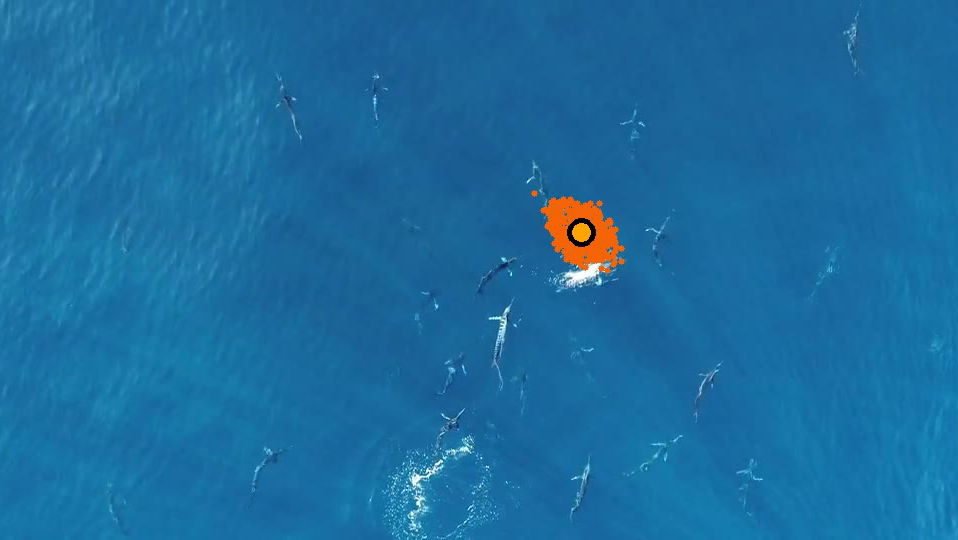}
\end{subfigure}\hspace{1em}
\begin{subfigure}[b]{.27\textwidth}
\centering\includegraphics[width=\textwidth]{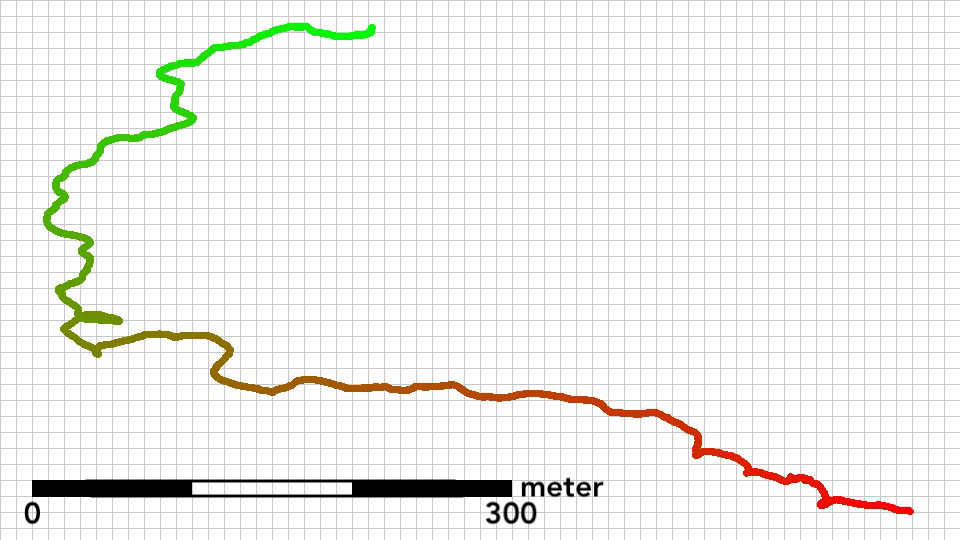}
\end{subfigure}\\
\begin{subfigure}[b]{.27\textwidth}
\centering\includegraphics[width=\textwidth]{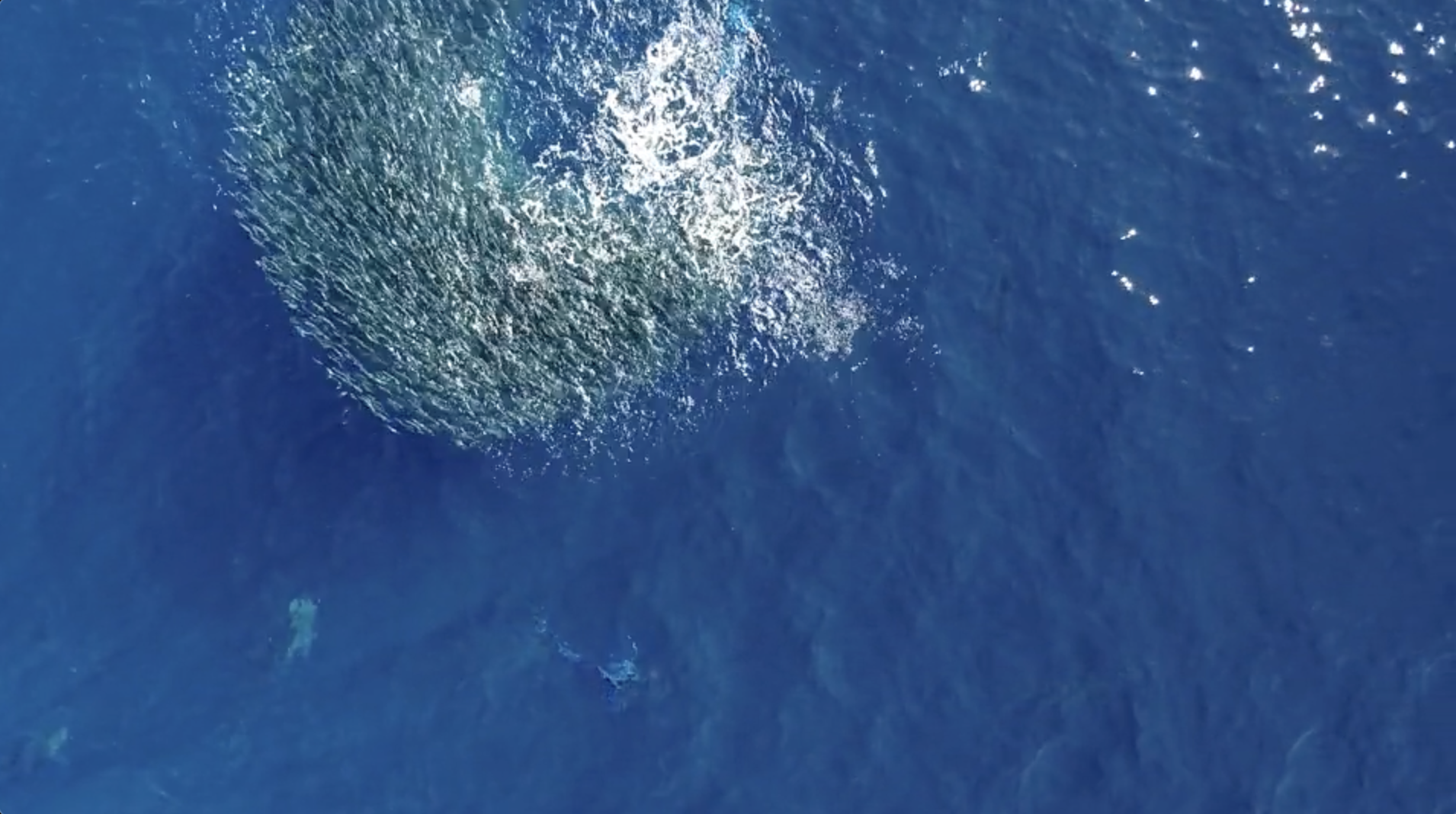}
\end{subfigure}\hspace{1em}\vspace{0.5em}
\begin{subfigure}[b]{.27\textwidth}
\centering\includegraphics[width=\textwidth]{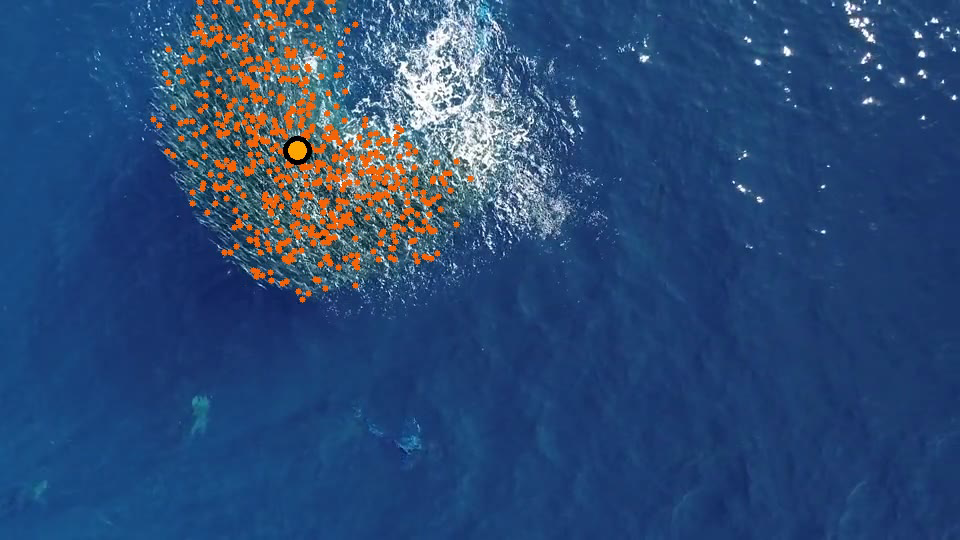}
\end{subfigure}\hspace{1em}
\begin{subfigure}[b]{.27\textwidth}
\centering\includegraphics[width=\textwidth]{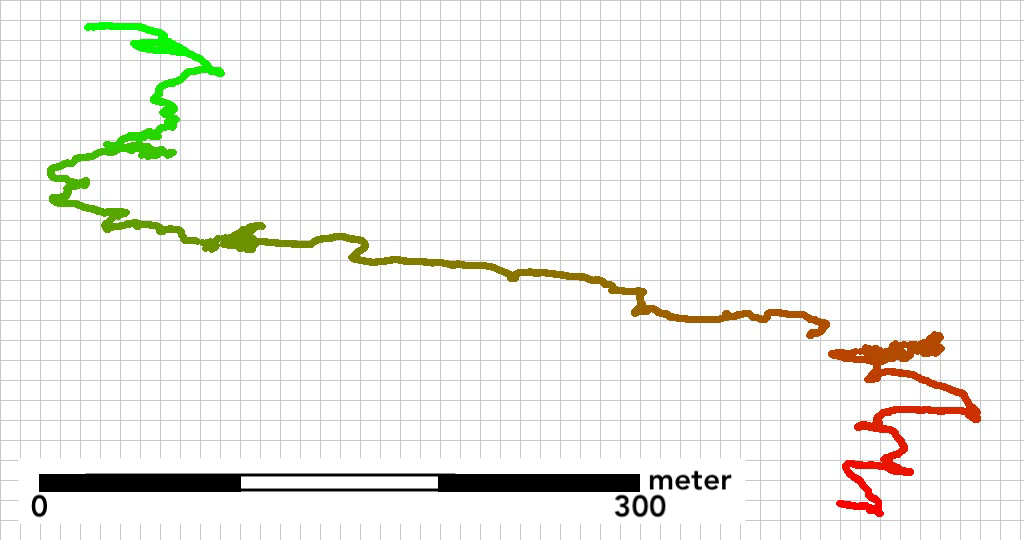}
\end{subfigure}\\
\begin{subfigure}[b]{.27\textwidth}
\centering\includegraphics[width=\textwidth]{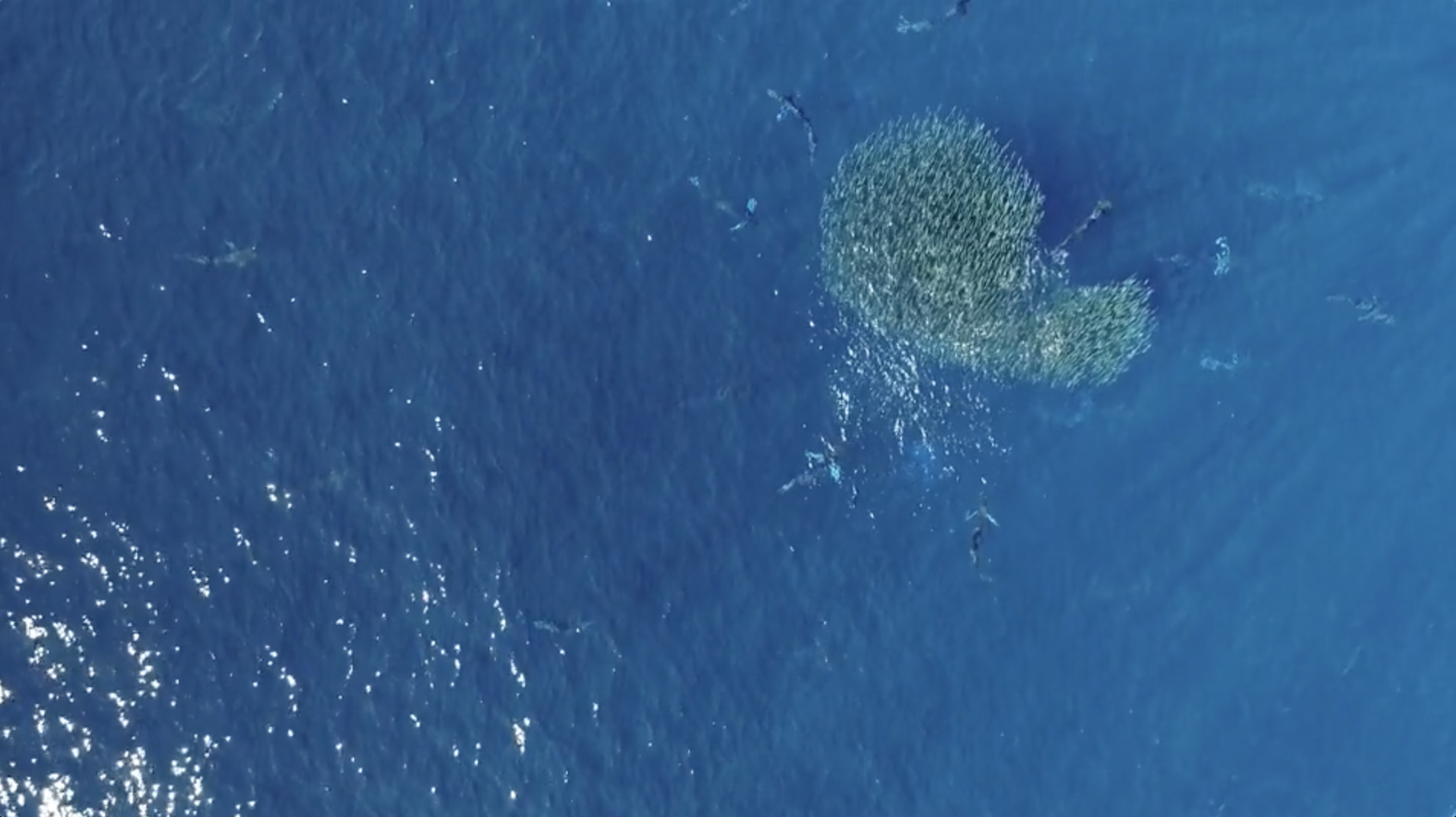}
\caption{Original frame}\label{fig:1f}
\end{subfigure}\hspace{1em}\vspace{0.5em}
\begin{subfigure}[b]{.27\textwidth}
\centering\includegraphics[width=\textwidth]{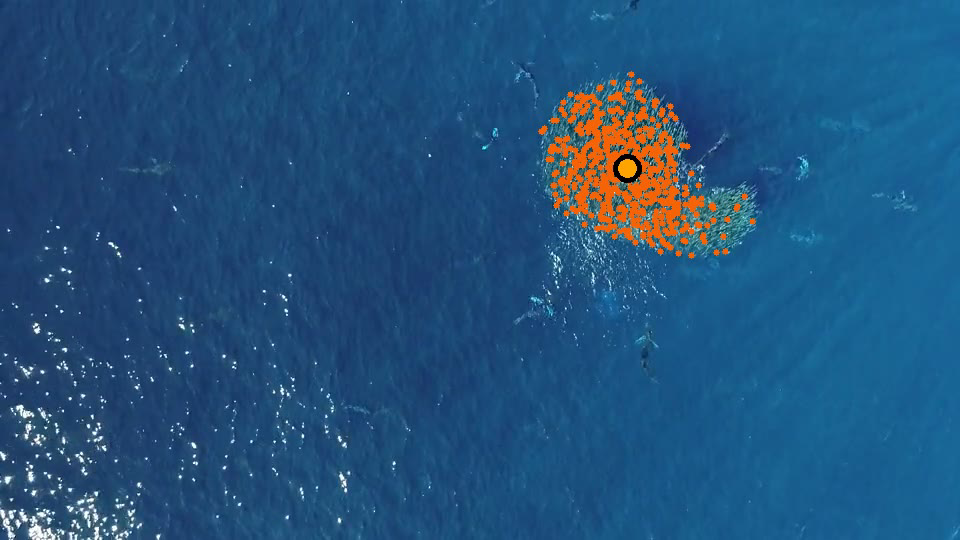}
\caption{SwDA - 2D particle tracking}\label{fig:1f}
\end{subfigure}\hspace{1em}
\begin{subfigure}[b]{.27\textwidth}
\centering\includegraphics[width=\textwidth]{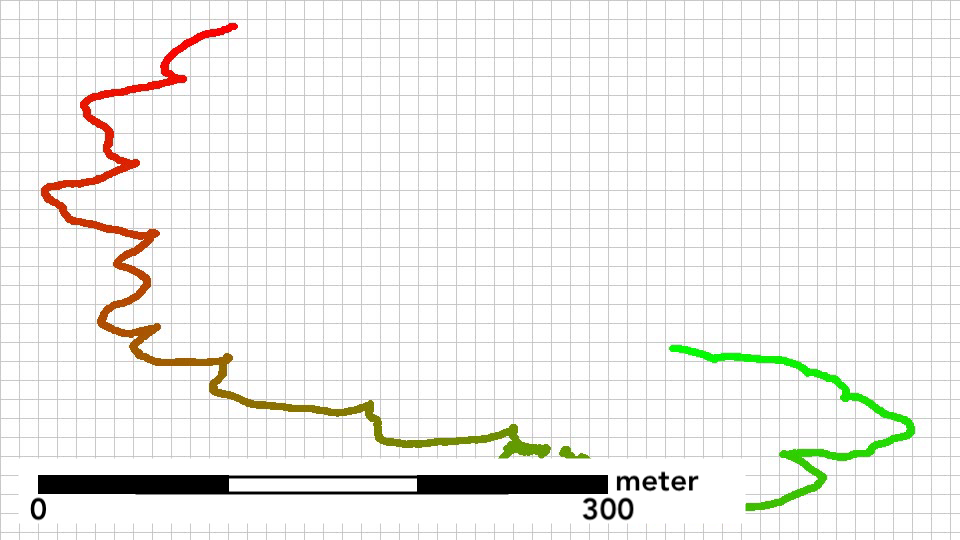}
\caption{3D motion trajectory (birds-eye view)}\label{fig:1f}
\end{subfigure}
\vspace{-0.5ex}
\caption{\textbf{Qualitative Results} of three different videos are shown: (a) Original frame, (b) Swarm detection via particle tracking and (c) Global movement trajectory of the swarm, 10m/grid cell, color visualises time. The overall video length is $\sim$5min. Full videos are provided in the accompanying suppl. material.}
\label{fig:results}
\vspace{-4mm}
\end{figure*}

{\bf Data.} Recordings picture highly dynamic predator and prey interactions in the Pacific Ocean 10-30 km offshore Baja California Sur. Data is captured by a DJI Phantom 4 pro drone at 60fps. Videos show a schooling prey during group hunts of striped marlins. Prey fish schools typically consist of populations ranging from approximately 100 to over 3000 individuals. Each video shows a different school of fish during predator attack. This wide range not only affects the school's appearance in terms of its size but also has a significant impact on the school's dynamics and behavior, especially in response to predatory attacks. Examples are shown in Figure~\ref{fig:results}. Each video file is accompanied with accurately synchronized sensor measurements of the drones on board sensors. In total 40min of video data is provided - 8 videos of 5min each. 
The videos are split into 4-folds. Per fold 4 videos are used for training, 2 videos are used for validation and testing respectively.
Per video 100 frames distributed over the full duration of each video are annotated with pixel accurate segmentation masks outlining the fish school's shape, resulting in a total of 800 segmentation masks - per k-fold 400 samples are used for training, 200 for validation and 200 for testing. The fish school's movement is tracked throughout the full video sequence with ground truth point annotations. Data will be made available for further analysis and research purposes.

\subsection{Evaluation}
This section evaluates and discusses multiple aspects of our tracking framework: estimation of movement trajectories, tracking of shape and localisation in world coordinates. 

\begin{figure}
    \centering
    \includegraphics[width=0.98\linewidth]{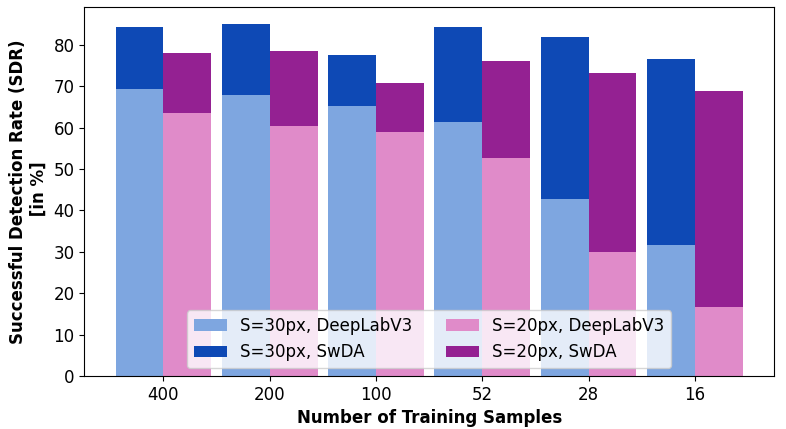}
    \vspace{-0.5ex}
    \caption{\textbf{Tracking Accuracy.} Evaluation of the Tracking Accuracy for different amount of labeled training data. Accuracy is measured via the successful detection rate (SDR).  Results for two different precision ranges are shown: SDR within a radius of 30 pixels (blue), SDR within a radius of 20 pixels (purple).}
    \label{fig:eval_traj}
\end{figure}

\begin{table}
\centering
\small
\begin{adjustbox}{max width=0.98\linewidth}
\setlength{\tabcolsep}{3pt}
\begin{tabular}{lccccccc}
\toprule 
\bf{Num Training Samples} &  & 400 &  &  & 16 & \\
\midrule 
\bf{Method} & $S=30$ & $S=20$ & $S=10$ & $S=30$ & $S=20$ & $S=10$\\
\midrule 
Follow Anything~\cite{maalouf2024follow} & 37.96 & 32.76 & 20.33 & 38.37 & 33.10 & 20.42 \\
DeepLabV3 & 69.31 & 63.65 & 45.06 & 31.63 & 16.81 & 6.61\\
\addlinespace[0.02cm]
SwDA & \bf 84.40 & \bf 77.93 & \bf 50.31 & \bf 76.66 & \bf 69.78 & \bf 40.22 \\
\bottomrule
\end{tabular}
\end{adjustbox}
\vspace{-0.5ex}
\caption{{\bf Tracking Accuracy.} 
We compare the quality of estimated 2D trajectories with \textit{Follow Anything}, a tracker that similarly tracks an object from a moving drone and \textit{DeepLabV3}. DeepLabV3 segments each frame in a sequence (such as a video) independently of the others, resulting in instantaneous segmentation masks. The mean point of each mask is tracked.}
\label{table:tracking-acc}
\end{table}

\begin{table}
\centering
\small
\begin{adjustbox}{max width=0.98\linewidth}
\setlength{\tabcolsep}{3pt}
\begin{tabular}{lccccc}
\toprule 
\bf{Method} & Intersection over Union & Precision & Recall & F1-measure \\
\midrule 
Follow Anything~\cite{maalouf2024follow} & 42.5 & 52.0 & 51.1 & 49.8 \\
DeepLabV3 & 73.8 & 83.1 & 77.4 & 78.9 \\
\addlinespace[0.02cm]
SwDA & 71.4 & 76.2 & 85.4 & 78.9\\
\bottomrule
\end{tabular}
\end{adjustbox}
\vspace{-0.5ex}
\caption{{\bf Shape Segmentation Accuracy.} 
To compare the quality of the swarm's shape, we convert the set of tracked particles to segmentation masks. Reconstructing the shape of a 2D point cloud on the plane is inferred through its corresponding $\alpha$-shape. 
}
\label{table:results-shape}
\vspace{-2ex}
\end{table}

{\bf Movement Trajectories} An often used measure to assess the 2D tracking accuracy is the average trajectory error~\cite{zhang2018tutorial}. A significant drawback of this measure however becomes apparent when the tracker loses the target, the output location can be random and the average error value may not measure the tracking performance correctly~\cite{wu2013online, babenko2010robust}. Instead, a widely accepted measure is the successful detection rate (SDR), a measure that reports the percentage of accurate detections within a predefined precision range. We compare our approach SwDA with Follow Anything and DeepLabV3. Follow Anything leverages foundation models like CLIP~\cite{radford2021learning}, DINO~\cite{caron2021emerging}, and SAM~\cite{kirillov2023segment} to compute segmentation masks that best align with the queried objects, therefore its performance is independent upon the amount of training data as one can see in Table~\ref{table:tracking-acc}. The queried object is identified using ground truth segmentation masks. DeepLabV3 relies frame-wise estimates - not exploiting temporal consistency. We follow a similar protocol as introduced in DeepLabCut~\cite{mathis2018deeplabcut} to extract trajectories from frame-wise segmentation masks. Fig~\ref{fig:eval_traj} shows the tracking accuracy averaged over all four training folds and for different amounts of training data. SwDA - our particle set tracker, shows significantly stronger tracking performance also in low-data regimes. While the accuracy of a pure learning based approach drops to 31.6\% and 38.4\% respectively (precision range of S=30px) when trained with only 16 training samples, the particle tracker only slightly reduces performance compared to training with all available training data. We improve upon
recent advancements in semantic object segmentation by integrating particle filter-based tracking. This approach leverages physical regularities such as temporal consistency commonly found in behavioral animal data.

{\bf Tracking of Collective Formation Patterns.} Shape segmentation accuracy is evaluated using four distinct accuracy metrics: intersection over union, precision, recall, and the F1-measure. In SwDA, soft segmentation masks from DeepLabV3 serve as frame-wise observations $o_t$, and particles are resampled accordingly. It is anticipated that this process will roughly maintain the segmentation quality expected from DeepLabV3. Given a set of particles, we approximate pixelwise segmentation masks by computing its corresponding $\alpha$-shape~\cite{edelsbrunner1983shape}. This leads to a slightly more spatially expanded segmentation regions, which can be seen in the higher recall of SwDA (Tab.~\ref{table:results-shape}).

{\bf 3D Tracking and Localization.} Based solely on maritime drone recordings without unique landmarks it is impossible to achieve a good estimate of the algorithms ability to extract real-world coordinates from 2D trajectories on the image plane. Therefor we record aerial video data capturing a simple terrestrial environment, with clear and easy to identify markers. The ground truth 3D position of markers are measured using real-time kinematic positioning (RTK). Eight markers are located at different positions and recorded by a moving drone. While the detection of the target position is meant to be kept as simple as possible, the goal of this experiment is to evaluate the algorithms ability to retrieve accurate relative distances between detected makers. The accuracy of reconstructing a 3D position of a static landmark is reported in Tab.~\ref{table:main-linear-velocity}. The drone's sensor measurements are fused to estimate the drone's pose.   

\begin{table}
\centering
\small
\begin{adjustbox}{max width=0.9\linewidth}
\setlength{\tabcolsep}{3pt}
\begin{tabular}{lccc}
\toprule 
\bf{Method} & absolute error [in m] & standard deviation \\
\midrule 
GPS & 0.41 & 0.28  \\
IMU & 0.96 & 0.65  \\
\addlinespace[0.02cm]
IMU+GPS (Kalman filter) & \bf 0.32 & \bf 0.21 \\
\bottomrule
\end{tabular}
\end{adjustbox}
\vspace{-0.5ex}
\caption{{\bf Tracking Accuracy in 3D.} 
We compare different localization approaches via GPS, IMU or their combination. SwDA integrates sensor measurements from IMU and GPS via Kalman filtering. The relative distances between markers is evaluated. Absolut distances of markers vary between 14m and 23m.}
\label{table:main-linear-velocity}
\vspace{-2ex}
\end{table}

\section{Conclusion}
This work unifies learning and probabilistic modeling within a coherent algorithmic structure. By exploiting temporal consistency which can be found in behavioral data, our proposed method can handle challenging training scenarios where not much annotated training data is available. Furthermore since tracking relies on both visual appearance features and measurements of physical sensors the proposed algorithm is capable to tracking animal behavior in highly demanding environments, such as the open ocean. 

\section*{Acknowledgements}
We thank Captain Marco and his crew at Magdalene Bay Whale Tours for assistance obtaining video of striped marlin. We thank Felipe Galván-Magaña from CICIMAR for logistical support. This project has been funded by the Deutsche Forschungsgemeinschaft (DFG, German Research Foundation) under Germany’s Excellence Strategy – EXC 2002/1 “Science of Intelligence” – project number 390523135. In addition, this work has been partially supported by MIAI@Grenoble Alpes, (ANR-19-P3IA-0003)

\section*{Ethics}
Field data was conducted under permits SGPA/DGVS/02460/18, SGPA/DGVS/01643/19 and SGPA/DGVS/08074/21, and we followed the ASAB ethics (Guidelines for the treatment of animals in behavioural research and teaching 2020, Animal Behaviour 159, i-xi) recommendations for fieldwork.
{
    \small
    \bibliographystyle{ieeenat_fullname}
    \bibliography{main}
}


\end{document}